\documentclass[11pt,letterpaper]{article}

\usepackage{naaclhlt2016}
\usepackage{color}
\usepackage{flushend}

\naaclfinalcopy 
\def\naaclpaperid{552} 

\usepackage{varwidth} 
\usepackage{url,graphicx,times}
\usepackage{tabularx,amsmath}
\usepackage{amssymb}
\usepackage{multirow}
\usepackage{booktabs}
\usepackage{enumerate}
\usepackage{subfigure}
\usepackage{enumitem}
\usepackage{setspace}

\date{}

\begin{document}

\title{Modeling Relational Information in Question-Answer Pairs with Convolutional Neural Networks}

\author{Aliaksei Severyn\thanks{\texttt{ }This work was carried out during his PhD in the University of Trento.} \\
Google Research\\
Zurich, Switzerland\\
\url{severyn@google.com}\\
\And Alessandro Moschitti  \\
 Qatar Computing Research Institute, HBKU\\
  5825 Doha, Qatar \\ 
 \url{amoschitti@qf.org.qa}
}

\maketitle

\begin{abstract}
\vspace{-.7em}
In this paper, we propose convolutional neural networks for learning an optimal representation of question and answer sentences. Their main aspect is the use of relational information given by the matches between words from the two members of the pair. The matches are encoded as embeddings with additional parameters (dimensions), which are tuned by the network. These allows for better capturing interactions between questions and answers, resulting in a significant boost in  accuracy. We test our models on two widely used answer sentence selection benchmarks. The results clearly show the effectiveness of our relational information, which allows our relatively simple network to approach the state of the art.
\end{abstract}


\vspace{-1.5em}
\section{Introduction}
\vspace{-.5em}
\label{sec:intro}
Modeling text pairs to compute their semantic similarity is at the core of many NLP tasks. 
%
The most common approach is to encode them with many complex lexical, syntactic and semantic features and then compute  various similarity measures between the obtained representations. 
%
%
Recently, it has been shown that the problem of semantic text matching can be tackled using distributional word matching, e.g., for matching questions with candidate answers~\cite{yih:acl2013}.

Deep learning approaches generalize the distributional word matching problem to matching sentences and take it one step further by learning the optimal sentence representations for a given task. 
Deep neural networks are able to effectively capture the compositional process of mapping the meaning of individual words in a sentence to a continuous representation of the sentence. In particular, it has been recently shown that convolutional neural networks are able to efficiently learn to embed input sentences into low-dimensional vector space preserving important syntactic and semantic aspects of the input sentence, which leads to state-of-the-art results in many NLP tasks~\cite{nal:acl2014,kim:2014:EMNLP2014,deep-qa}. 


In this paper, we capitalize on our previous work \cite{severyn2015learning} extending it with a novel deep learning architecture for modelling question-answer pairs for answer sentence reranking.  The main building blocks of our architecture are two distributional sentence models based on convolutional neural networks  (ConvNets). These underlying sentence models work in parallel, mapping questions and answer sentences to their distributional vectors, which are then used to learn the semantic similarity between them. 
To compute question-answer similarity score we adopt an approach used in the deep learning model of~\cite{deep-qa}, which produces excellent results on the answer sentence selection task. 
However, their model only operates on unigram or bigrams, while our architecture learns to extract and compose n-grams of higher degrees, thus allowing for capturing longer range dependencies. Additionally, our architecture uses not only the intermediate representations of questions and answers to compute their similarity but also includes them in the final representation, which constitutes a much richer representation of the question-answer pairs. 

The main novelty of our architecture is the way we choose to model relational information in a pair. Yu et al.~\shortcite{deep-qa} combine the output of their deep learning model with additional features in the final logistic regression model. Such features count the number word overlaps between the two pair members. This provides a sort of relational information, which significantly improves the network accuracy. 
In contrast, our model uses a completely different approach, which injects relational information about matching words directly into the word embeddings as additional dimensions. The augmented word embeddings are thus passed through the layers of the convolutional feature extractors: this enables the automatic encoding of the relations between question-answer pairs in a more structured manner. Moreover, our embedding dimensions encoding matches are parameters of the network and are tuned during the training.

In summary, the distinctive properties of our model are: (i) we use a state-of-the-art distributional sentence model for learning to map input sentences to vectors, which are then used to measure the similarity between them; (ii) our model encodes question-answer pairs in a richer representation using not only their similarity score but also their intermediate representations; (iii) we augment the word embeddings with additional dimensions to encode the fact that certain words overlap in a given question-answer pair and let the network tune these parameters; (iv)  the architecture of our network makes it straightforward to include any additional features encoding question-answer similarities; and finally (v) our model is trained end-to-end starting from the input sentences to producing a final score that is used to rerank answers. We only require to initialize word embeddings trained on some large unsupervised corpora. However, given a large training set the network can also optimize the embeddings directly for the task, thus omitting the need for pre-training of the word embeddings.

We test our model on a popular answer sentence selection benchmark TREC13 \cite{wang:2007} and on the more recent dataset  WikiQA \cite{yang-yih-meek:2015:EMNLP}. The results show the importance of using relational information and on WikiQA our network reaches the state of the art, i.e., an MRR of 71.07 and an MAP of 69.51.




\vspace{-.7em}
\section{Our Deep Learning Model}
\vspace{-.7em}

This section explains the architecture of our deep learning model for modelling question-answer pairs to rerank answer sentences. 
We treat the answer sentence selection problem as a simple binary classification where answer candidates with higher prediction scores are ranked above the ones with lower scores. More formally, each question $\mathbf{q}_i \in Q$ is associated with a list of labelled candidate answer sentences $\{(y_{i1}, \mathbf{a}_{i1}), \dots, (y_{in}, \mathbf{a}_{in}) \}$, where labels  $y_{ij} \in \{0,1\}$ with $1$ corresponding to answers that contain a correct answer and $0$ otherwise. Our goal is to learn a decision function that maps each question-answer pair to a score reflecting their similarity: $f(\mathbf{\theta},\psi(\mathbf{q}_i, \mathbf{a}_{ij}))$, where $\psi(\cdot)$ is a function encoding question-answer pairs into a joint feature space, and $\mathbf{\theta}$ are model parameters. 

Given that we choose to model the answer sentence selection task as a binary classification, our main effort lies in designing a deep learning architecture for learning an optimal representation of question-answer pairs.
Its main building blocks are two distributional \textit{sentence models} based on convolutional neural networks. These underlying sentence models work in parallel mapping question and answer sentences to their distributional vectors, which are then used to learn the similarity between them. 

In the following, we first describe our \textit{sentence model} for mapping queries and documents to their intermediate representations and then describe how they can be used for learning semantic matching between input query-document pairs. 

\vspace{-.5em}
\subsection{Distributional sentence model}
\begin{figure}[t]
\center
\includegraphics[width=\linewidth]{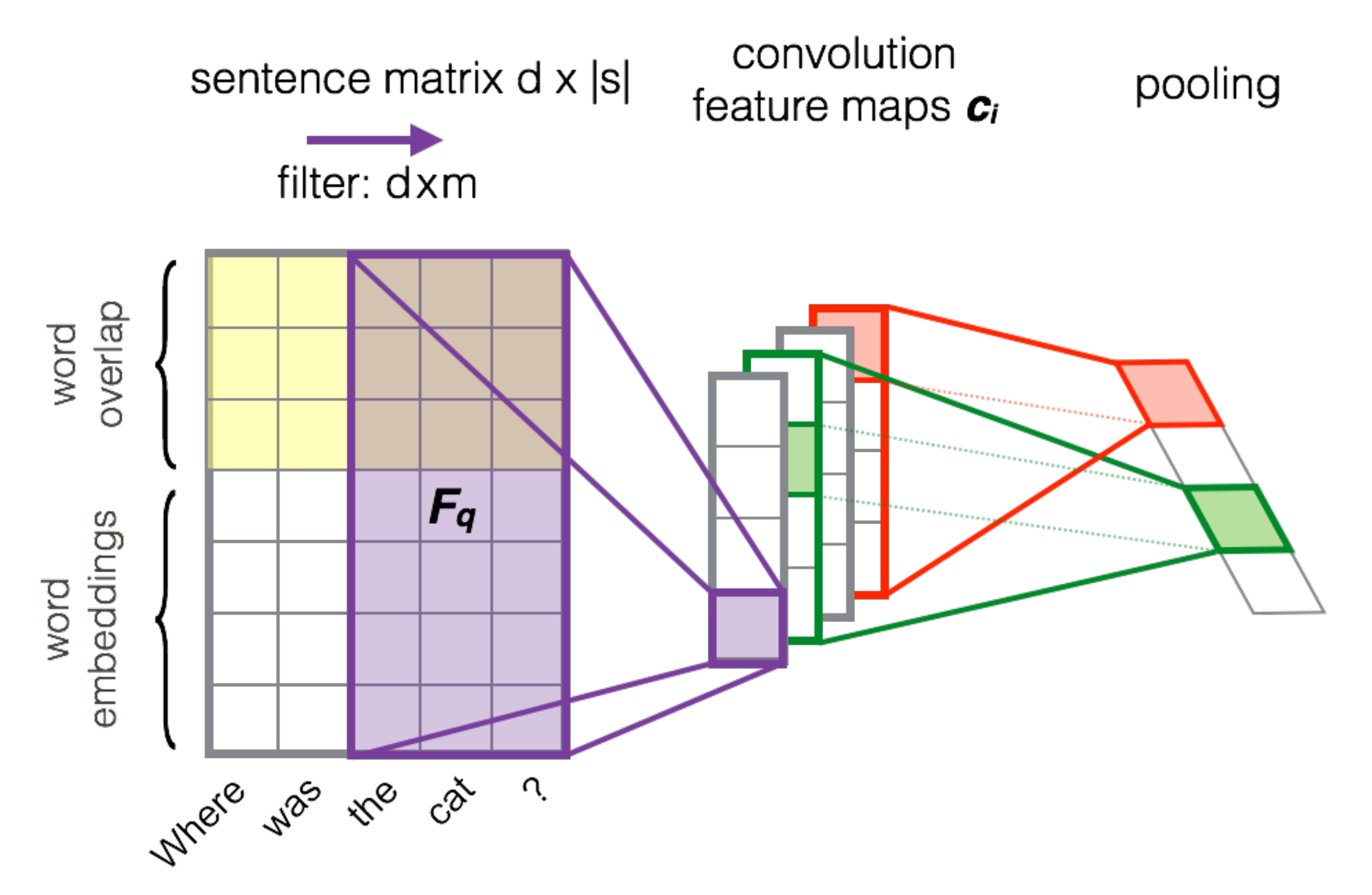}
\vspace{-2em}
\caption{Our sentence model for mapping input sentences to their intermediate representations. 
}
\vspace{-1em}
\label{nnet-sentence}
\end{figure}

\vspace{-.3em}
\label{convnets}
The architecture of our network for mapping sentences to feature vectors is shown on Fig.~\ref{nnet-sentence}.  It is mainly inspired by the convolutional architectures used in \cite{nal:acl2014,kim:2014:EMNLP2014} for performing various sentence classification tasks. However, different from previous work the goal of our distributional sentence model is to learn intermediate representations of questions and answers used to compute their semantic matching.

\begin{figure*}[t]
\center
\includegraphics[width=0.8\linewidth]{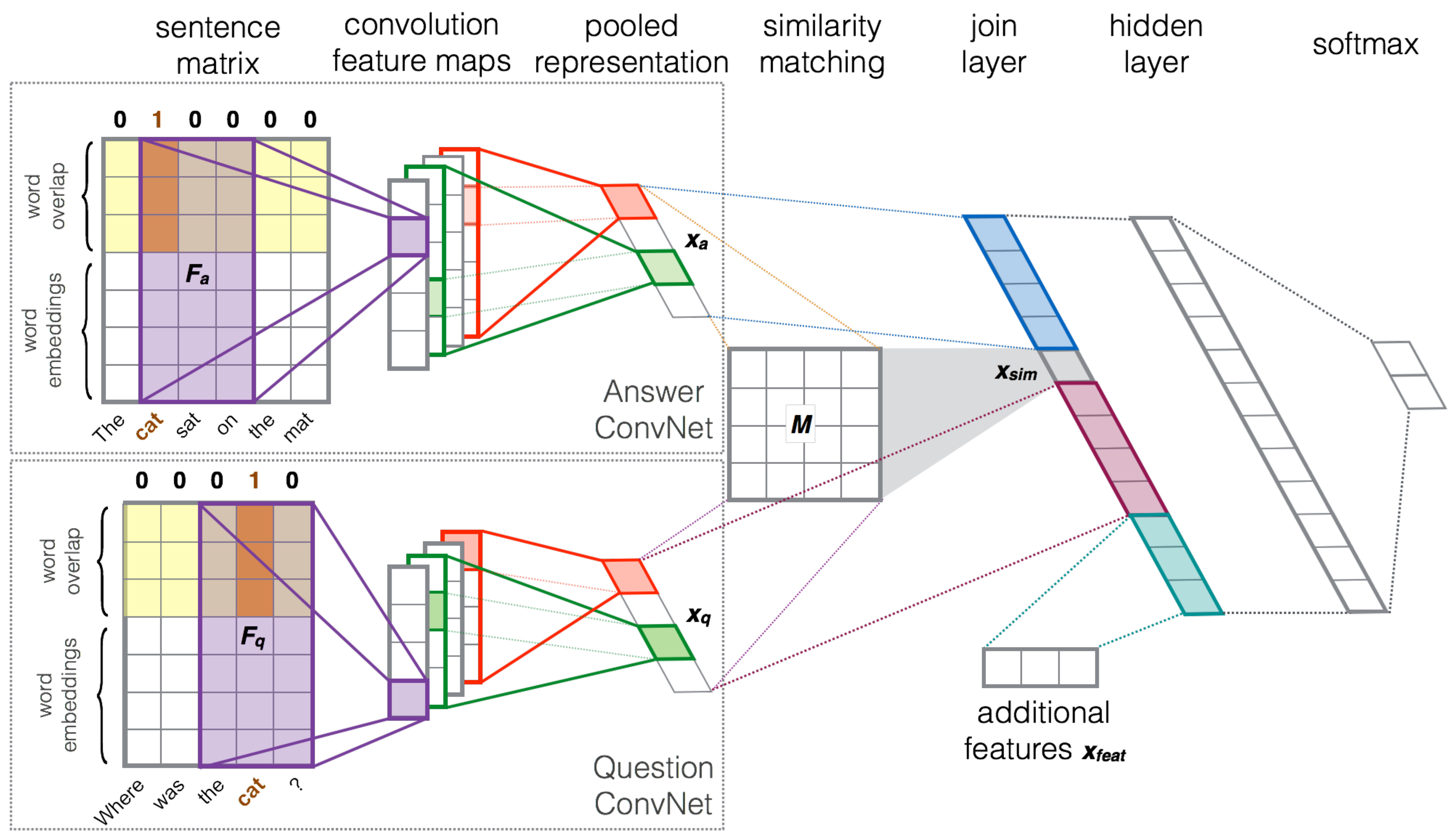}
\caption{Our deep learning architecture for reranking question-answer pairs. The relational information in a pair is modelled by augmenting word embeddings with additional dimensions to encode overlapping words, e.g., we feed the network with additional \textit{word overlap} indicator features whose values equal to 1 correspond to words that overlap in a pair, e.g., a non-stop word \texttt{cat}.}
\label{network}
\vspace{-1em}
\end{figure*}

Our sentence model is composed of a single \textit{wide} convolutional layer followed by a non-linearity and simple \textit{max} pooling.
In the following we give a brief explanation of its main components: sentence matrix, activations, convolutional and pooling layers.

\vspace{-.3em}
\subsubsection{Sentence matrix}
\vspace{-.2em}
\label{sec:word-overlap}
The input to the network are raw words that need to be translated into real-valued feature vectors to be processed by subsequent layers of the network. 

The input is a sentence $\mathbf{s}$ treated as a sequence of words: ${[w_1, \dots, w_{|s|}]}$, where each word is drawn from a finite-sized vocabulary $V$.  
The architecture of neural networks is not well suited for dealing with discrete words, hence, they are represented by low-dimensional, real-valued, dense vectors $\mathbf{w} \in \mathbb{R}^{d_w}$ looked up in a matrix $\mathbf{W} \in \mathbb{R}^{d_w \times |V|}$ (whose columns correspond to words in $V$). The mapping from words to their word embeddings is performed by a lookup table operation $LT_{\mathbf{W}}(w_i)=\mathbf{w}_i$. 
Hence, for each input sentence $\mathbf{s}$ we build a sentence matrix $\mathbf{S}$ where each $i$-th column corresponds to a word embedding $\mathbf{w}_i$.

To learn to capture and compose features of individual words in a given sentence from low-level word embeddings into higher level semantic concepts, the neural network applies a series of transformations to the input sentence matrix $\mathbf{S}$ using convolution, non-linearity and pooling operations, which we describe next. 

\vspace{-.3em}
\subsubsection{Convolutional feature maps}
\vspace{-.2em}
The aim of the convolutional layer is to extract patterns, i.e., discriminative word sequences that are common throughout the training instances.

More formally, the convolution operation $*$ between an input matrix $\mathbf{S} \in \mathbb{R}^{d \times |s|}$ and a filter (or a convolution kernel)  $\mathbf{F} \in \mathbb{R}^{d \times m}$ of width $m$  results in a vector $\mathbf{c} \in \mathbb{R}^{|s|+m-1}$ where each component is computed as follows:\vspace{-.5em}
\begin{equation}
\mathbf{c}_i = (\mathbf{S} \ast \mathbf{F})_i = \sum_{k,j} (\mathbf{S}_{[:,i-m+1:i]} \otimes \mathbf{F})_{kj}
\label{eq:convolution}
\vspace{-.7em}
\end{equation}
where $\otimes$ is the element-wise multiplication and $\mathbf{S}_{[:,i-m+1:i]}$ is a matrix slice of size $m$ along the columns. Note that the convolution filter is of the same dimensionality $d$ as the input sentence matrix. As shown in Fig.~\ref{nnet-sentence}, it slides along the column dimension of $\mathbf{S}$ producing a vector $\mathbf{c} \in \mathbb{R}^{|s|-m+1}$ in output. Each component $c_i$ is the result of computing an element-wise product between a column slice of $\mathbf{S}$ and a filter matrix $\mathbf{F}$, which is then summed to a single value.

So far we have described a way to compute a convolution between the input sentence matrix and a single filter.
To form a richer representation of the data, deep learning models apply a set of filters that work in parallel generating multiple feature maps (also shown on Fig.~\ref{nnet-sentence}). A set of filters form a filter bank $\mathbf{F}\in \mathbb{R}^{n \times d \times m}$ sequentially convolved with the sentence matrix $\mathbf{S}$ and producing a feature map matrix $\mathbf{C} \in \mathbb{R}^{n \times (|s|-m+1)}$. 

In practice, we also need to add a bias vector $\mathbf{b} \in \mathbb{R}^{n}$ to the result of a convolution -- a single $b_i$ value for each feature map $\mathbf{c}_i$. This allows the network to learn an appropriate threshold.

\vspace{-.3em}
\paragraph{Activation units.}
\vspace{-.3em}
To enable the learning of non-linear decision boundaries, each convolutional layer is typically followed by a non-linear activation function $\alpha()$ applied element-wise. 
We use a rectified linear (ReLU) function defined as simply $max(0, \mathbf{x})$ in our model since, as shown in~\cite{hinton:icml2010}, it speeds up the training and sometimes produces more accurate results.

\vspace{-.3em}
\paragraph{Pooling.}
\vspace{-.3em}
The output from the convolutional layer (passed through the activation function) are then passed to the pooling layer, whose goal is to aggregate the information and reduce the representation. 
%
We use \textit{max} pooling in our model which simply returns the maximum value. It operates on columns of the feature map matrix $\mathbf{C}$ returning the largest value: $\text{pool}(\mathbf{c}_i): \mathbb{R}^{|s|+m-1} \rightarrow \mathbb{R}$ (also  shown schematically in Fig.~\ref{nnet-sentence}).


Convolutional layer passed through the activation function together with pooling layer acts as a non-linear feature extractor. Given that multiple feature maps are used in parallel to process the input, deep learning networks are able to build rich feature representations of the input. 

This ends the description of our \textit{sentence} model. In the following we present our deep learning architecture for learning to match short text pairs.

\vspace{-.5em}
\subsection{Our relational model}
\vspace{-.3em}
When learning to match text pairs, modelling the relational connections between sentences has been shown to greatly improve the accuracy of the semantic similarity models. For example, top performing systems on Semantic Textual Similarity benchmarks~\cite{sts:2015} rely on similarity scores obtained by aligning the words and phrases between sentences in a pair. Yih et al.~\shortcite{yih:acl2013} also uses latent word-alignment structure in their semantic similarity model to compute similarity between question and answer sentences. Yu et al.~\shortcite{deep-qa} achieves large improvements by combining the output of their deep learning model with word count features in a  logistic regression model.

To allow our convolutional neural network capture the connections between related words in a pair we feed it with an additional binary-like input about overlapping words.  
In particular, for each word $w$ in the input sentence we associate an additional \textit{word overlap} indicator feature $o \in \{0, 1\}$, where $1$ corresponds to words that overlap in a given pair and $0$ otherwise (see Fig.~\ref{network}). To decide if the words overlap, we perform string matching.  

Hence, we require an additional lookup table layer for the word overlap features $LT_{\mathbf{W}_o}(\cdot)$  with parameters $\mathbf{W}_o \in \mathbb{R} ^{d_o \times 2}$, where $d_o \in \mathbb{N}$ is the number of dimensions to encode word overlap features and is a hyper-parameter of the model. Effectively, we are augmenting word embeddings with additional dimensions that encode the fact that a given word in a pair is overlapping or semantically similar and let the network learn its optimal representation.
Given a word $w_i$ its, final word embedding $\mathbf{w}_i \in \mathbb{R}^d$ (where $d = d_w + d_o$) is obtained by concatenating the output of two lookup table operations $LT_{\mathbf{W}}(w_i)$ and $LT_{\mathbf{W_o}}(w_i)$ (also see Fig.~\ref{network}).

\vspace{-.2em}
\subsection{The matching model}
\vspace{-.2em}
The architecture of our model for matching question-answer pairs is presented in Fig.~\ref{network}.
Our sentence models (described in Sec.~\ref{convnets}) learn to map input sentences to vectors, which can then be used to compute their similarity. These are then used to compute a similarity score, which together with the distributional vector of question and answer sentences are used in a single joint representation.

In the following we describe how the intermediate representations produced by the sentence model can be used to compute question-answer similarity scores and give a brief explanation of the remaining layers, e.g. hidden and softmax.

\vspace{-.3em}
\paragraph{Similarity model.}
\vspace{-.3em}
Given the output of our sentence models, their resulting vector representations $\mathbf{x}_q$ and  $\mathbf{x}_a$, can be used to compute a question-answer similarity score. We follow the approach of~\cite{bordes:qa-matching} that defines the similarity between $\mathbf{x}_{q}$ and $\mathbf{x}_a$ vectors as follows:
\vspace{-.5em}
\begin{equation}
sim(\mathbf{x}_q, \mathbf{x}_a) = \mathbf{x}_q^T \mathbf{M} \mathbf{x}_a,
\label{qa-matching}
\vspace{-.8em}
\end{equation}
where $\mathbf{M} \in \mathbb{R}^{d \times d}$ is a similarity matrix. Eq.~\ref{qa-matching} can be viewed as a model of the noisy channel approach from machine translation, which has been widely used as a scoring model in information retrieval and QA~\cite{noisy-channel}. In this model, we seek a transformation of the candidate document $\mathbf{x}_a'=\mathbf{M}\mathbf{x}_a$ that is the closest to the input query $\mathbf{x}_q$. The similarity matrix $\mathbf{M}$ is a parameter of the network and is optimized during the training.

\vspace{-.3em}
\paragraph{Hidden and classification layers.}
\vspace{-.3em}
Our model includes an additional hidden layer right before the softmax layer (described next) to allow for modeling interactions between the components of the intermediate representation.  It computes the following transformation:
$
\alpha(\mathbf{w}_{h} \cdot \mathbf{x} + b),
$
where $\mathbf{w}_{h}$ is the weight vector of the hidden layer and $\alpha()$ is non-linearity.
Finally, to transform the output of the network to the probability distribution over the labels we apply a softmax function.


\vspace{-.5em}
\subsection{The information flow}
\vspace{-.2em}

The output of our sentence models (Sec.~\ref{convnets}) are distributional representations of a question $\mathbf{x}_q$ and an answer $\mathbf{x}_a$, which are then matched using a similarity matrix $\mathbf{M}$ according to Eq.~\ref{qa-matching}. This produces a single score $x_{\text{sim}}$ capturing various aspects of similarity (syntactic and semantic) of the question-answer pair. 
Note that it is also straight-forward to add additional features $\mathbf{x}_{\text{feat}}$ to the model. 

The join layer concatenates all intermediate vectors, the similarity score and any additional features into a single vector: 
$
\mathbf{x}_{\text{join}} = [\mathbf{x}_q^T; x_\text{sim}; \mathbf{x}_a^T; \mathbf{x}_{\text{feat}}^T]
$
This vector is then passed  through a fully connected hidden layer, which allows for modelling interactions between the components of the joined representation vector. Finally, the output of the hidden layer is further fed to the softmax classification layer, which generates a distribution over the class labels.

\vspace{-.5em}
\subsection{Training}
\vspace{-.3em}
The model is trained to minimise the negative conditional log-likelihood of the training set:
\vspace{-.4em}
\begin{equation*}
\begin{array}{ll}
\mathcal{C} & = -\text{log} \prod_{i=1}^N p(y_i|\mathbf{q}_i, \mathbf{a}_i; \mathbf{\theta})
\end{array}
\vspace{-.4em}
\end{equation*}
where 
$\theta$ contains all the network parameters: 
\vspace{-.5em}
\[
\mathbf{\theta} = \{ \mathbf{W}; \mathbf{W}_o; \mathbf{F}_q; \mathbf{b}_q; \mathbf{F}_a; \mathbf{b}_a; \mathbf{M}; \mathbf{w}_h; b_h; \mathbf{w}_s; b_s\},
\vspace{-.5em}
\] 
namely the word embeddings matrix $\mathbf{W}$ and word overlap feature matrix $\mathbf{W}_o$, filter weights and biases of the convolutional layers, similarity matrix $\mathbf{M}$, parameters of the hidden and softmax layers. 

The parameters of the network are optimized by stochastic gradient descent (SGD) using backpropogation algorithm to compute the gradients. 





\section{Experiments and Evaluation}
This section describes the dataset and our experimental setup, also giving details about how we obtain the word embeddings matrix $\mathbf{W}$ and train our network.

\vspace{-.5em}
\subsection{Data and setup}
\vspace{-.4em}
We test our model on the manually curated TREC QA dataset\footnote{\url{http://cs.stanford.edu/people/mengqiu/data/qg-emnlp07-data.tgz}} from ~\newcite{wang:2007}, which appears to be one of the most widely used benchmarks for answer sentence  reranking. 
The set of questions are collected from TREC QA tracks 8-13. The manual judgement of candidate answer sentences is provided for the entire TREC 13 set and for the first 100 questions from TREC 8-12. 

To enable a direct comparison with the previous work, we use the same \texttt{train}, \texttt{dev} and \texttt{test} sets. Table~\ref{table:qa-data} summarizes the datasets used in our experiments. An additional training set TRAIN-ALL provided by \newcite{wang:2007}  contains 1,229 questions from the entire TREC 8-12 collection and comes with automatic judgements. This set represents a more noisy setting, nevertheless, it provides many more QA pairs for learning. 

\noindent{\textbf{WikiQA.}} This is an the open domain QA  dataset \cite{yang-yih-meek:2015:EMNLP}. Its questions were derived from the Bing query logs and the candidate answers were extracted from paragraphs of the associated Wikipedia pages. The training, test, and development set contain 2,118, 633 and 296 questions, respectively. Consistently with~\cite{yin2015abcnn}, we remove the questions without answers for our evaluations.

\noindent{\textbf{Evaluation.}}
The two metrics used to evaluate the quality of our model are Mean Average Precision (MAP) and Mean Reciprocal Rank (MRR).
We use the official \texttt{trec\_eval} scorer to compute the above metrics.

\begin{table}[!t]
\vspace{-.4em}
\small
\centering
\caption{TREC QA datasets for answer reranking.
}
\label{table:qa-data}
\begin{tabular}{lrrr}
\toprule
Data & \# Questions & \# QA pairs & \% Correct\\
\midrule
TRAIN-ALL & 1,229 & 53,417 & 12.0\% \\
TRAIN & 94 & 4,718 & 7.4\%  \\
DEV & 82 & 1,148 & 19.3\%  \\
TEST & 100 & 1,517 & 18.7\%  \\
\bottomrule
\end{tabular}
\vspace{-1em}
\end{table}

\label{sec:word-emb}
\noindent{\textbf{Word vectors.}}
While our model allows for learning the word embeddings directly, we keep the word matrix parameter $\mathbf{W}$ static. This is due to a common experience that a minimal size of the dataset required for tuning the word embeddings for a given task should be at least in the order of hundred thousands, while in our case the number of question-answer pairs is one order of magnitude smaller. Hence, similar to~\cite{Denil2014b,kim:2014:EMNLP2014,deep-qa} we keep the word embeddings fixed and initialize the word matrix $\mathbf{W}$ from an unsupervised neural language model. 

We run \texttt{word2vec} tool~\cite{word2vec} on the English Wikipedia dump and the AQUAINT corpus\footnote{\url{https://catalog.ldc.upenn.edu/LDC2002T31}} containing roughly 375 million words. We opt for a skipgram model with window size 5 and filtering words with frequency less than 5. We set the dimensionality of our word embeddings to 50 to be on the line with~\cite{deep-qa}. The resulting model contains 50-dimensional vectors for about 3.5 million words. Embeddings for words  not present in the \texttt{word2vec} model are randomly initialized with each component uniformly sampled. 

We minimally preprocess the data only performing tokenization and lowercasing all words. To reduce the size of the resulting vocabulary $V$, we also replace all digits with 0. 
The size of the word vocabulary $V$ for experiments using TRAIN set is 17,023 with approximately 95\% of words initialized using \texttt{wor2vec} embeddings and the remaining 5\% words are initialized at random. For the TRAIN-ALL setting the $|V|=56,953$ with about 90\% words found in the \texttt{word2vec} model.
\vspace{-.7em}
\paragraph{Word matching features.}
In contrast to the word embeddings matrix, the size of the vocabulary $V_o$ to encode word overlap features is tiny. Given such a small parameter space it is possible to tune the vectors even on small sized datasets. Hence, we keep this as a parameter optimized by our network.  We set the size of the space, $d_o$, to 5 and randomly initialize the entries of the matrix $\mathbf{W}^o$ by sampling from the uniform distribution.



\vspace{-.7em}
\subsection{Training and hyperparameters}
\vspace{-.5em}
The parameters of our deep learning model were chosen on a \texttt{dev} set: the width $m$ of the convolution filters is 5 and the number of convolutional feature maps is 100.  We use ReLU activation function and a simple max-pooling. 
The size of the hidden layer is equal to the size of the $\mathbf{x}_{\text{join}}$ vector obtained after concatenating question and answer sentence vectors from the distributional models, similarity score and additional features.

To train the network we use stochastic gradient descent with shuffled mini-batches. We eliminate the need to tune the learning rate by using the Adadelta update rule~\cite{adadelta}. The batch size is set to 50 examples. The network is trained for 25 epochs with early stopping, i.e., we stop the training if no update to the best accuracy on the \texttt{dev} set has been made for the last 5 epochs. The accuracy computed on the \texttt{dev} set is the MAP score. At test time we use the parameters of the network obtained with the best MAP score on the \texttt{dev} set: we compute the MAP score after each 10 mini-batch updates and save the network parameters if a new best \texttt{dev} score is obtained. In practice, the training converges after a few epochs.

%
%
%


\vspace{-.5em}
\subsection{Results and discussion}
\vspace{-.3em}
Our goal is to evaluate the impact of using our: (i) more powerful convolutional network for sentence modeling; (ii) distributional representations of questions and answers in addition to the similarity score; and (iii) approach to model matching words by augmenting word embeddings with additional dimensions vs. providing the network with a pre-computed feature vector of overlapping word counts as in~\cite{deep-qa}.

\subsubsection{Distributional sentence models}
Table~\ref{table:qa-exp} summarises the results for the setting when the network is trained using only input question-answer pairs without using any additional features, i.e., we omit the word overlap features. 

First, we report the results of our model when using only a similarity score $x_{\text{sim}}$. 
It should be noted that the network by Yu et al.~\shortcite{deep-qa}, similarly to ours, relies on a convolutional neural network to learn intermediate representations. However, their convolutional neural network operates only on unigram or bigrams, while in our architecture we use a larger width of the convolution filter, thus allowing for capturing longer range dependencies. 

Additionally, along with the question-answer similarity score from Eq.~\ref{qa-matching}, our architecture includes intermediate representations of the question and the answer $\mathbf{x}_q$ and $\mathbf{x}_a$ into the final vector representation $\mathbf{x}_{\text{join}}$, which together constitute a much richer representation for computing the final score. We call this network simply CNN.


\begin{table}[t]
\centering
\small
\caption{Results on TREC QA using only similarity score (Sim) and also including the distributional representation (Dist) of question and answer sentences.}
\label{table:qa-exp}
\vspace{0.5em}
\begin{tabular}{|c|c|c|c|c|}
\hline
		& \multicolumn{2}{c|}{Train} & \multicolumn{2}{c|}{Train-ALL}\\
\hline
\hline
	 	& Sim		& Dist 		& Sim		& Dist 		\\
\hline
MAP		& .5884		& .6258		& .6521		& .6709		\\
\hline
MRR		& .6036		& .6591		& .7010		& .7280		\\
\hline
\end{tabular}
\vspace{-.5em}
\end{table}


\vspace{-.5em}
\subsubsection{Relational models}
\vspace{-.2em}
Yu et al.~\shortcite{deep-qa} shows that combining the output of their deep learning system with a simple feature vector that includes word overlap counts in a logistic regression model, provides a significant boost in accuracy and yields new state-of-the-art results. 

Table~\ref{table:qa-exp-count} provides the results when we include the information about overlapping words in two modes: (i) feature vector (\textit{fvec}) mode -- when we include overlapping word counts replicating~\cite{deep-qa}, which is represented by a feature vector $\mathbf{x}_{\text{feat}}$ that is plugged into the final representation $\mathbf{x}_{\text{join}}$ (see Fig.~\ref{network}); and (ii) \textit{embeddings} mode -- when we augment the representation of input words with additional word \textit{overlap} indicator features (as described in Sec.~\ref{sec:word-overlap})\footnote{a combined model using both \textit{fvec} and \textit{embeddings} modes yielded the same performance as using the \textit{embeddings} model.}.
First, we note that the results are significantly better than in Table~\ref{table:qa-exp} when no overlap information is used. Adding word overlap information in the form of a feature vector $\mathbf{x}_{\text{feat}}$ results in a considerable generalization improvement of the network. As argued by Yu et al.~\shortcite{deep-qa}, distributional word embeddings have certain shortcomings especially when dealing with proper nouns and cardinal numbers, which are frequent in factoid questions.

In contrast, our approach to encode the relational information about overlapping words in a pair (\textit{embeddings}) directly into word embeddings shows even larger improvement on TRAIN, achieving the best results on TRAIN-ALL with a MAP score of 76.54\% and an MRR of 81.86\%. We call this network using relational information, CNN$_{R}$.




\begin{table}[!t]
\centering
\small
\caption{Results on TREC QA when augmenting the deep learning model with relational information about overlapping words.}
\label{table:qa-exp-count}
\vspace{0.5em}
\begin{tabular}{|c|c|c|c|c|}
\hline
		& \multicolumn{2}{c|}{Train} & \multicolumn{2}{c|}{Train-ALL}\\
\hline
\hline
	 	& fvec		& Emb. 		& fvec		& Emb. 		\\
\hline
MAP		& .7275		& .7325		& .7459		& .7654		\\
\hline
MRR		&  .7796		& .8018		& .8078		& .8186		\\
\hline
\end{tabular}
\end{table}


\begin{table}[t]
\centering
\small
\caption{Survey of the results on the TREC QA answer selection task (after score rescaling).}
\label{table:qa-exp-soa}
\vspace{0.5em}
\begin{tabular}{lrrr}
\toprule
Model & MAP & MRR \\
\midrule
Wang et al. (2007) & .6029 & .6852 \\
Heilman and Smith (2010) & .6091 & .6917 \\
Wang and Manning (2010) & .5951 & .6951 \\
Yao et al. (2013) & .6307 & .7477 \\
Severyn \& Moschitti (2013) & .6781 & .7358 \\
Yih et al. (2013)  & .7092 & .7700 \\
Yu et al. (2014) & .7113 & .7846 \\
\newcite{wang2015faq} 	& .7063 & .7740  \\ 
\newcite{yin2015abcnn} 	& .6951 & .7633  \\
\newcite{miao2015neural}& \textbf{.7339} & \textbf{.8117}  \\\hline

\midrule
CNN$_{R}$ on (TRAIN) & .6857 & {.7660} \\
CNN$_{R}$ on (TRAIN-ALL) & .7186 & {.7828} \\
\bottomrule
\end{tabular}
\end{table}

\begin{table}[t]
\vspace{-1em}
\center
\small
\begin{tabular}{|p{4cm}|r|r|r|}
\hline
&MAP&MRR&P@1\\
\hline
\hline
\multicolumn{4}{|c|}{\textbf{State of the art}}\\
\hline
CNN$_{c}$~\cite{yang-yih-meek:2015:EMNLP}		&.6520&.6652&n/a\\
ABCNN~\cite{yin2015abcnn}					&.6914&\textbf{.7127}&n/a\\
LSTM$_{a,c}$~\cite{miao2015neural}			&.6855&.7041&n/a\\
NASM$_{c}$~\cite{miao2015neural}				&.6886&.7069&n/a\\ 
\hline
\hline
\multicolumn{4}{|c|}{\textbf{Our Models}}\\
\hline
CNN 		&.6661	& .6851	&.5401\\
CNN$_R$		&\textbf{.6951}	& .7107	&.5720\\
\hline
\hline
\end{tabular}

\caption{Performance on the WikiQA dataset}
\vspace{-1em}
\label{tab:wikiqatest}
\end{table}

\vspace{-.1em}
\subsection{Comparing with the state of the art}
\vspace{-.1em}
It should be noted that, to be consistent with the results of previous work on TREC13, it is required to evaluate our models in the same setting as~\cite{wang:2007,yih:2014}, i.e., we need to (i) remove the questions having only correct or only incorrect  answer sentence candidates and (ii) use the same evaluation script and the gold judgment file as they used. As pointed out by Footnote 7 in~\cite{yih:2014}, the evaluation script always considers 4 questions to be answered incorrectly thus penalizing the overall score of the systems.
This basically lowered the performance of CNN$_{R}$ from an MAP and an MRR of .7654 and .8186  to .7186 and .7828, respectively. We used these numbers in Table~\ref{table:qa-exp-soa} for exactly comparing with the results of previously published systems.
We note that our model is almost on par with previous models. However, TREC13 is too small to assess the rank of our approach.
Therefore, we evaluated our best systems on WikiQA, which being larger, enables a more reliable system comparison. 
Tab. \ref{tab:wikiqatest} reports the system performance on WikiQA. It shows that our model reaches the accuracy of the best system, ABCNN, and outperforms NASM$_{c}$ by \newcite{miao2015neural}, which was superior to our models on TREC13.
Finally, the difference between CNN and CNN$_{R}$ is again remarkable confirming the benefit of relational information.



\vspace{-.5em}
\section{Related Work}
\vspace{-.5em}
\label{sec:related-work}
Most of the previous work to tackle the answer sentence selection task use various approaches to model transformations of syntactic trees between a question and its candidate answer sentence, e.g., Wang et al.~\shortcite{wang:2007} use quasi-synchronous grammar, Heilman \& Smith~\shortcite{heilman:naacl:2010} develop an improved Tree Edit Distance (TED) model,  Wang \& Manning ~\shortcite{wang_manning:acl:2010} develop a probabilistic model to learn tree-edit operations on dependency parse trees, while Yao et al.~\shortcite{yao:naacl:2013} applies linear chain CRFs with features derived from TED.
Severyn and Moschitti~\shortcite{severyn-moschitti:2013:EMNLP} applied SVM with tree kernels to shallow syntactic representations. Yih et al.~\shortcite{yih:acl2013} use distributional models based on lexical semantics to match semantic relations of aligned words in QA pairs. 

Recently, deep learning approaches have been successfully applied to various sentence classification tasks, e.g.,~\cite{nal:acl2014,kim:2014:EMNLP2014}, and for modelling text pairs, e.g.~\cite{dnn-matching:nips:2013,dnn-matching:nips:2014}, where in the latter model they use up to 3 convolution-pooling layers, while in our experiments deeper architectures were severely overfitting and we compensate our more shallow sentence ConvNets by using a more powerful relational model.

Additionally, a number of deep learning models have been recently applied to question answering, e.g., Yih et al.~\shortcite{yih:2014} applied convolutional neural networks to  open-domain question answering; Bordes et al.~\shortcite{bordes:qa-matching} propose a neural embedding model combined with the knowledge base for open-domain QA; Iyyer et al.~\shortcite{Iyyer:2014} applied recursive neural networks to the factoid QA over paragraphs.

The work closest to ours is \cite{deep-qa}, where they present a deep learning architecture for answer sentence selection. However, their sentence model to map questions and answers to vectors operates only on unigrams or bigrams. Our sentence model is based on a convolutional neural network that uses a larger width of the convolution filter, thus allowing the network to capture longer range dependencies. Moreover, our architecture along with the similarity score also encodes vector representations of questions and answers used to compute the final score. Hence, our model constructs and learns a richer representation of the question-answer pairs, which results in superior results on the answer sentence selection dataset. Moreover, we use a completely different way to encode relational information about words that overlap in a pair. Finally, our deep learning model is trained end-to-end, while in~\cite{deep-qa} they use the output of their neural network in a separate logistic scoring model.



\vspace{-.5em}
\section{Conclusions and future work}
\vspace{-.5em}
\label{sec:conclusions}

In this paper, we propose a novel deep learning architecture for answer sentence selection. 
Our experimental findings show that our model can achieve the accuracy of state-of-the-art networks, which are much more complex. 
This is largely due to our use of more expressive models for the input question and answer sentences, and our approach to inject relational information directly in the word embeddings. However, our word overlap indicator features are based on simple string matching, which is clearly a very coarse way to model relatedness between words in a question-answer pair.

Recently, deep learning architectures have been successfully applied to learn word alignments in machine translation, e.g.,~\cite{dnn:word-alignment}. It sounds promising to allow the network to learn to dynamically align the related words in a question and its answer. This in turn requires to maximize over the latent alignment configurations, thus making the optimization problem highly non-convex. Our preliminary experiments show that a far larger number of text pairs are required to train such architectures. We leave it for the future work.

\bibliographystyle{naaclhlt2016}
\bibliography{sigir2015,acl2015}

\begin{thebibliography}{}

\bibitem[\protect\citename{Agirre \bgroup et al.\egroup }2015]{sts:2015}
Eneko Agirre, Carmen Banea, Claire Cardie, Daniel Cer, Mona Diab, Aitor
  Gonzalez-Agirre, Weiwei Guo, Iñigo Lopez-Gazpio, Montse Maritxalar, Rada
  Mihalcea, German Rigau, Larraitz Uria, and Janyce Wiebe.
\newblock 2015.
\newblock {SemEval-2015 Task 2: Semantic Textual Similarity, English, Spanish
  and Pilot on Interpretability}.
\newblock In {\em Proceedings of the 9th International Workshop on Semantic
  Evaluation (SemEval 2015)}.

\bibitem[\protect\citename{Bordes \bgroup et al.\egroup
  }2014]{bordes:qa-matching}
Antoine Bordes, Jason Weston, and Nicolas Usunier.
\newblock 2014.
\newblock Open question answering with weakly supervised embedding models.
\newblock In {\em ECML}.

\bibitem[\protect\citename{Denil \bgroup et al.\egroup }2014]{Denil2014b}
Misha Denil, Alban Demiraj, Nal Kalchbrenner, Phil Blunsom, and Nando
  de~Freitas.
\newblock 2014.
\newblock Modelling, visualising and summarising documents with a single
  convolutional neural network.
\newblock Technical report, University of Oxford.

\bibitem[\protect\citename{Echihabi and Marcu}2003]{noisy-channel}
Abdessamad Echihabi and Daniel Marcu.
\newblock 2003.
\newblock A noisy-channel approach to question answering.
\newblock In {\em ACL}.

\bibitem[\protect\citename{Heilman and Smith}2010]{heilman:naacl:2010}
Michael Heilman and Noah~A. Smith.
\newblock 2010.
\newblock Tree edit models for recognizing textual entailments, paraphrases,
  and answers to questions.
\newblock In {\em NAACL}.

\bibitem[\protect\citename{Hu \bgroup et al.\egroup
  }2014]{dnn-matching:nips:2014}
Baotian Hu, Zhengdong Lu, Hang Li, and Qingcai Chen.
\newblock 2014.
\newblock Convolutional neural network architectures for matching natural
  language sentences.
\newblock In {\em NIPS}.

\bibitem[\protect\citename{Iyyer \bgroup et al.\egroup }2014]{Iyyer:2014}
Mohit Iyyer, Jordan Boyd-Graber, Leonardo Claudino, Richard Socher, and Hal
  {Daum\'e III}.
\newblock 2014.
\newblock A neural network for factoid question answering over paragraphs.
\newblock In {\em EMNLP}.

\bibitem[\protect\citename{Kalchbrenner \bgroup et al.\egroup
  }2014]{nal:acl2014}
Nal Kalchbrenner, Edward Grefenstette, and Phil Blunsom.
\newblock 2014.
\newblock A convolutional neural network for modelling sentences.
\newblock {\em ACL}.

\bibitem[\protect\citename{Kim}2014]{kim:2014:EMNLP2014}
Yoon Kim.
\newblock 2014.
\newblock Convolutional neural networks for sentence classification.
\newblock In {\em EMNLP}, pages 1746--1751, Doha, Qatar, October.

\bibitem[\protect\citename{Lu and Li}2013]{dnn-matching:nips:2013}
Zhengdong Lu and Hang Li.
\newblock 2013.
\newblock A deep architecture for matching short texts.
\newblock In {\em NIPS}.

\bibitem[\protect\citename{Miao \bgroup et al.\egroup }2015]{miao2015neural}
Yishu Miao, Lei Yu, and Phil Blunsom.
\newblock 2015.
\newblock Neural variational inference for text processing.
\newblock {\em arXiv preprint arXiv:1511.06038}.

\bibitem[\protect\citename{Mikolov \bgroup et al.\egroup }2013]{word2vec}
Tomas Mikolov, Ilya Sutskever, Kai Chen, Greg~S Corrado, and Jeff Dean.
\newblock 2013.
\newblock Distributed representations of words and phrases and their
  compositionality.
\newblock In {\em Advances in Neural Information Processing Systems 26}, pages
  3111--3119.

\bibitem[\protect\citename{Nair and Hinton}2010]{hinton:icml2010}
Vinod Nair and Geoffrey~E. Hinton.
\newblock 2010.
\newblock Rectified linear units improve restricted boltzmann machines.
\newblock In {\em ICML}.

\bibitem[\protect\citename{Severyn and
  Moschitti}2013]{severyn-moschitti:2013:EMNLP}
Aliaksei Severyn and Alessandro Moschitti.
\newblock 2013.
\newblock Automatic feature engineering for answer selection and extraction.
\newblock In {\em EMNLP}.

\bibitem[\protect\citename{Severyn and Moschitti}2015]{severyn2015learning}
Aliaksei Severyn and Alessandro Moschitti.
\newblock 2015.
\newblock Learning to rank short text pairs with convolutional deep neural
  networks.
\newblock In {\em Proceedings of the 38th International ACM SIGIR Conference on
  Research and Development in Information Retrieval}, pages 373--382. ACM.

\bibitem[\protect\citename{Wang and Ittycheriah}2015]{wang2015faq}
Zhiguo Wang and Abraham Ittycheriah.
\newblock 2015.
\newblock Faq-based question answering via word alignment.
\newblock {\em arXiv preprint arXiv:1507.02628}.

\bibitem[\protect\citename{Wang and Manning}2010]{wang_manning:acl:2010}
Mengqiu Wang and Christopher~D. Manning.
\newblock 2010.
\newblock Probabilistic tree-edit models with structured latent variables for
  textual entailment and question answer- ing.
\newblock In {\em ACL}.

\bibitem[\protect\citename{Wang \bgroup et al.\egroup }2007]{wang:2007}
Mengqiu Wang, Noah~A. Smith, and Teruko Mitaura.
\newblock 2007.
\newblock What is the jeopardy model? a quasi-synchronous grammar for qa.
\newblock In {\em EMNLP}.

\bibitem[\protect\citename{Xuchen~Yao and Callison-Burch}2013]{yao:naacl:2013}
Peter~Clark Xuchen~Yao, Benjamin Van~Durme and Chris Callison-Burch.
\newblock 2013.
\newblock Answer extraction as sequence tagging with tree edit distance.
\newblock In {\em NAACL}.

\bibitem[\protect\citename{Yang \bgroup et al.\egroup
  }2013]{dnn:word-alignment}
Nan Yang, Shujie Liu, Mu~Li, Ming Zhou, and Nenghai Yu.
\newblock 2013.
\newblock Word alignment modeling with context dependent deep neural network.
\newblock In {\em ACL}.

\bibitem[\protect\citename{Yang \bgroup et al.\egroup
  }2015]{yang-yih-meek:2015:EMNLP}
Yi~Yang, Wen-tau Yih, and Christopher Meek.
\newblock 2015.
\newblock Wikiqa: A challenge dataset for open-domain question answering.
\newblock In {\em Proceedings of the 2015 Conference on Empirical Methods in
  Natural Language Processing}, pages 2013--2018, Lisbon, Portugal, September.
  Association for Computational Linguistics.

\bibitem[\protect\citename{Yih \bgroup et al.\egroup }2013]{yih:acl2013}
Wen-Tau Yih, Ming-Wei Chang, Christopher Meek, and Andrzej Pastusiak.
\newblock 2013.
\newblock Question answering using enhanced lexical semantic models.
\newblock In {\em ACL}, August.

\bibitem[\protect\citename{Yih \bgroup et al.\egroup }2014]{yih:2014}
Wen-Tau Yih, Xiaodong He, and Christopher Meek.
\newblock 2014.
\newblock Semantic parsing for single-relation question answering.
\newblock In {\em ACL}.

\bibitem[\protect\citename{Yin \bgroup et al.\egroup }2015]{yin2015abcnn}
Wenpeng Yin, Hinrich Sch{\"u}tze, Bing Xiang, and Bowen Zhou.
\newblock 2015.
\newblock Abcnn: Attention-based convolutional neural network for modeling
  sentence pairs.
\newblock {\em arXiv preprint arXiv:1512.05193}.

\bibitem[\protect\citename{Yu \bgroup et al.\egroup }2014]{deep-qa}
Lei Yu, Karl~Moritz Hermann, Phil Blunsom, and Stephen Pulman.
\newblock 2014.
\newblock Deep learning for answer sentence selection.
\newblock {\em CoRR}.

\bibitem[\protect\citename{Zeiler}2012]{adadelta}
Matthew~D. Zeiler.
\newblock 2012.
\newblock Adadelta: An adaptive learning rate method.
\newblock {\em CoRR}.

\end{thebibliography}

\end{document}